\documentclass[10pt,twocolumn,letterpaper]{article}
\pdfoutput=1

\usepackage{cvpr}
\usepackage{times}
\usepackage{epsfig}
\usepackage{graphicx}
\usepackage{amsmath}
\usepackage{amssymb}
\usepackage{subcaption}
\usepackage{multicol}
\usepackage{enumitem}
\usepackage{amsfonts}
\usepackage{siunitx}
\usepackage{verbatim}
\usepackage{authblk}

\usepackage[breaklinks=true,bookmarks=false]{hyperref}

\cvprfinalcopy 

\begin{document}

\title{ArbiText: Arbitrary-Oriented Text Detection in Unconstrained Scene}
\author{Daitao Xing$^{1}$$^{,}$$^{3}$}
\author{Zichen Li$^{1}$$^{,}$$^{3}$}
\author{Xin Chen$^{4}$}
\author{Yi Fang$^{1}$$^{,}$$^{2}$$^{,}$$^{3}$\thanks{5 Metrotech Center LC024, Brooklyn, NY, 11201; Email:  yfang@nyu.edu; Tel:~+1-646-854-8866}}
\affil{$^{1}$NYU Multimedia and Visual Computing Lab \\ $^{2}$NYU Abu Dhabi, UAE \\ $^{3}$New York University, Brooklyn, NY, USA \\  $^{4}$HERE Technologies}

\maketitle

\begin{abstract}
Arbitrary-oriented text detection in the wild is a very challenging task, due to the aspect ratio, scale, orientation, and illumination variations. In this paper, we propose a novel method, namely Arbitrary-oriented Text (or ArbText for short) detector, for efficient text detection in unconstrained natural scene images. Specifically, we first adopt the circle anchors rather than the rectangular ones to represent bounding boxes, which is more robust to orientation variations. Subsequently, we incorporate a pyramid pooling module into the Single Shot MultiBox Detector framework, in order to simultaneously explore the local and global visual information, which can therefore generate more confidential detection results. Experiments on established scene-text datasets, such as the ICDAR 2015 and MSRA-TD500 datasets, have demonstrated the superior performance of the proposed method, compared to the state-of-the-art approaches. 
\end{abstract}

\section{Introduction}
\noindent Understanding texts in the wild plays an important role in many real-world applications such as PhotoOCR \cite{bissacco_photoocr:_2013}, road sign detection in intelligent vehicles \cite{chen_traffic_2011}, license plate detection \cite{masood_license_2017}, and assistive technology for the visually impaired \cite{ezaki_text_2004} \cite{bastan_mt3s:_2016}. To achieve this goal, the task of accurate arbitrary-oriented text detection becomes extremely important. Conventionally, when dealing with horizontal texts under controlled environments, this task can be accomplished through character-based methods such as \cite{huang_robust_2014}, \cite{jaderberg_deep_2014}, \cite{neumann_real-time_2016}, and \cite{wang_end--end_2012} considering that individual letters can be easily segmented and distinguished. However, in an unconstrained natural-scene image, text detection becomes rather challenging due to uncontrolled text variations and uncertainties, such as multi-orientation, text distortion, background noise, occlusion, and illumination changes. To address these problems, a lot of recent efforts have been devoted to employing state-of-the-art generic object detectors, such as the Fully Convolutional Network(FCN) \cite{shelhamer_fully_2016}, Region-based Convolutional Neural Network(R-CNN) \cite{ren_faster_2015}, and Single Shot Detector(SSD) \cite{liu_ssd:_2016}, for the purpose of text detection in the wild.

\begin{figure}
\begin{multicols}{2}
\includegraphics[width=4.2CM]{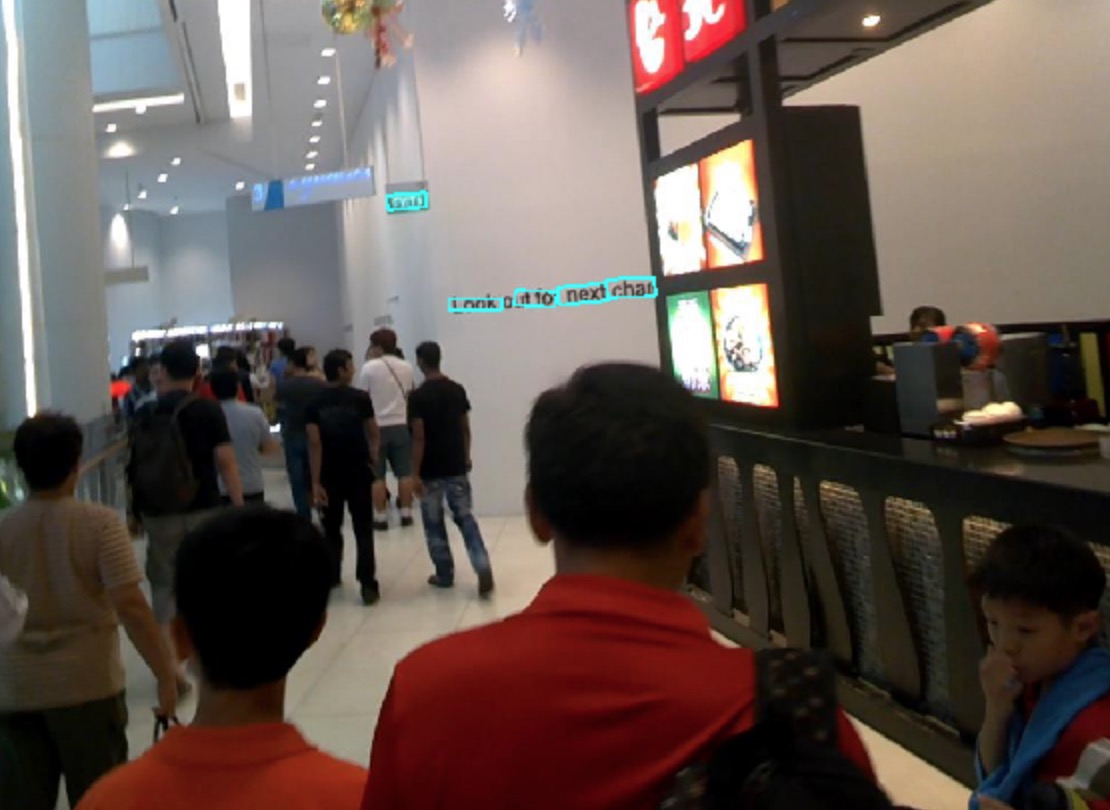}\par 
\includegraphics[width=4.2CM]{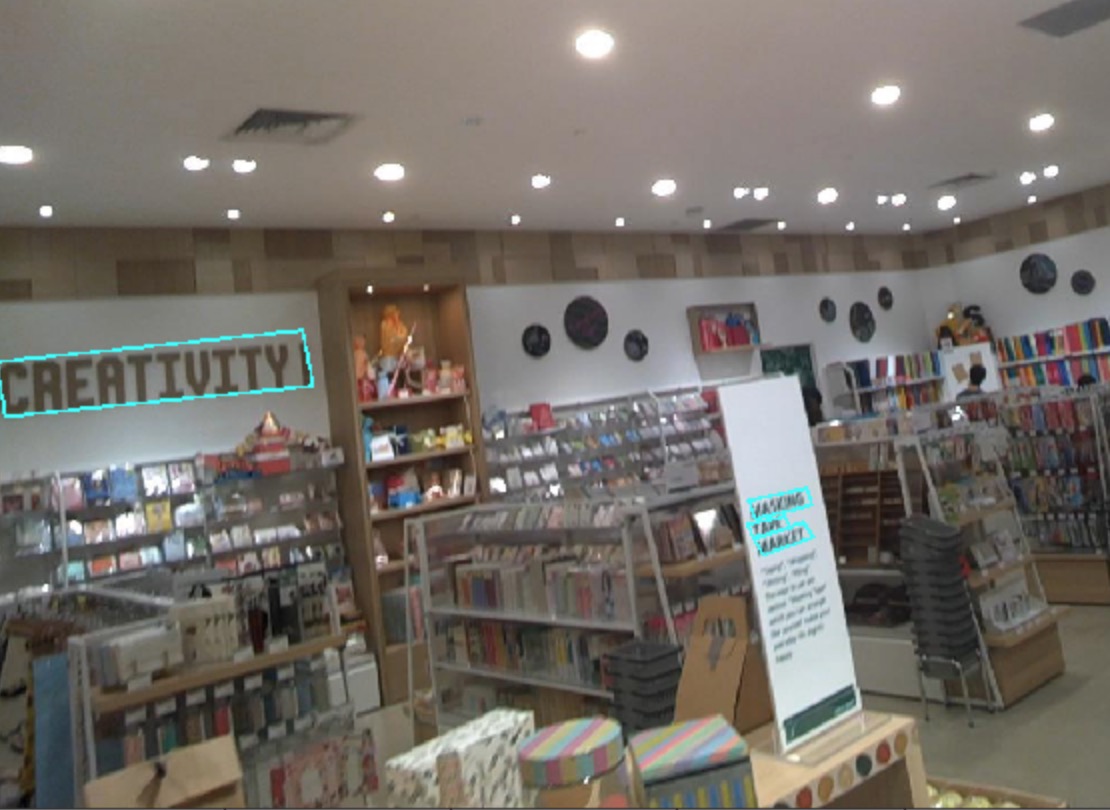}\par 
\end{multicols}
\begin{multicols}{2}
\includegraphics[width=4.2CM]{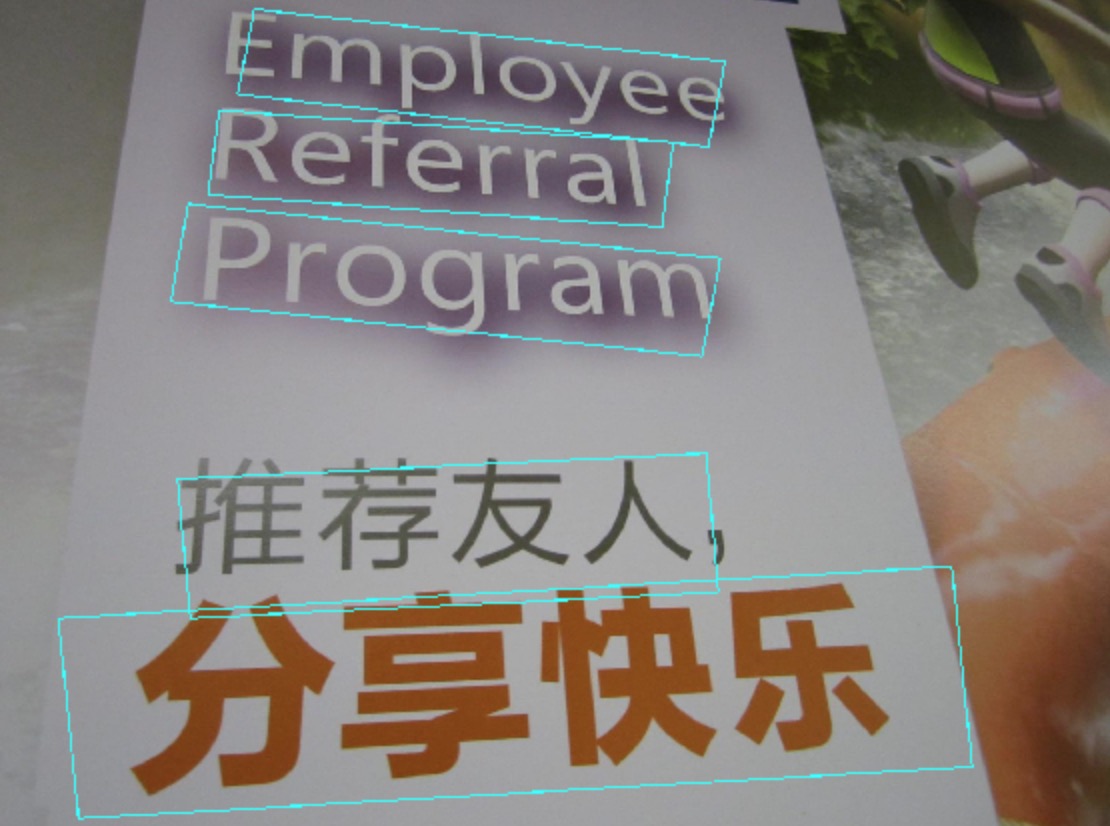}\par 
\includegraphics[width=4CM]{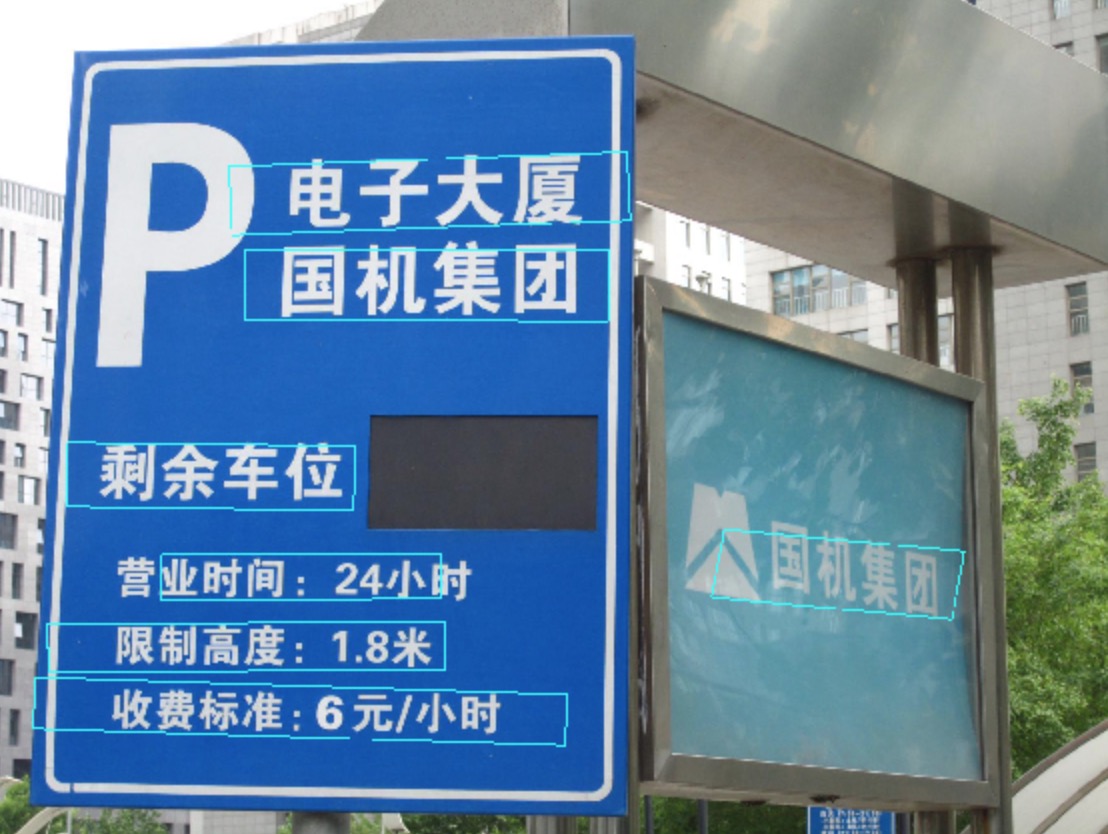}\par 
\end{multicols}
\caption{Text detection results of the proposed \textbf{ArbiText} method. Images in the first row shows examples from ICDAR2015 dataset, the second row shows examples MSRA-TD500 dataset}
\label{fig:1}
\end{figure}

\begin{figure*}[t]
\centering
\includegraphics[width=\textwidth]{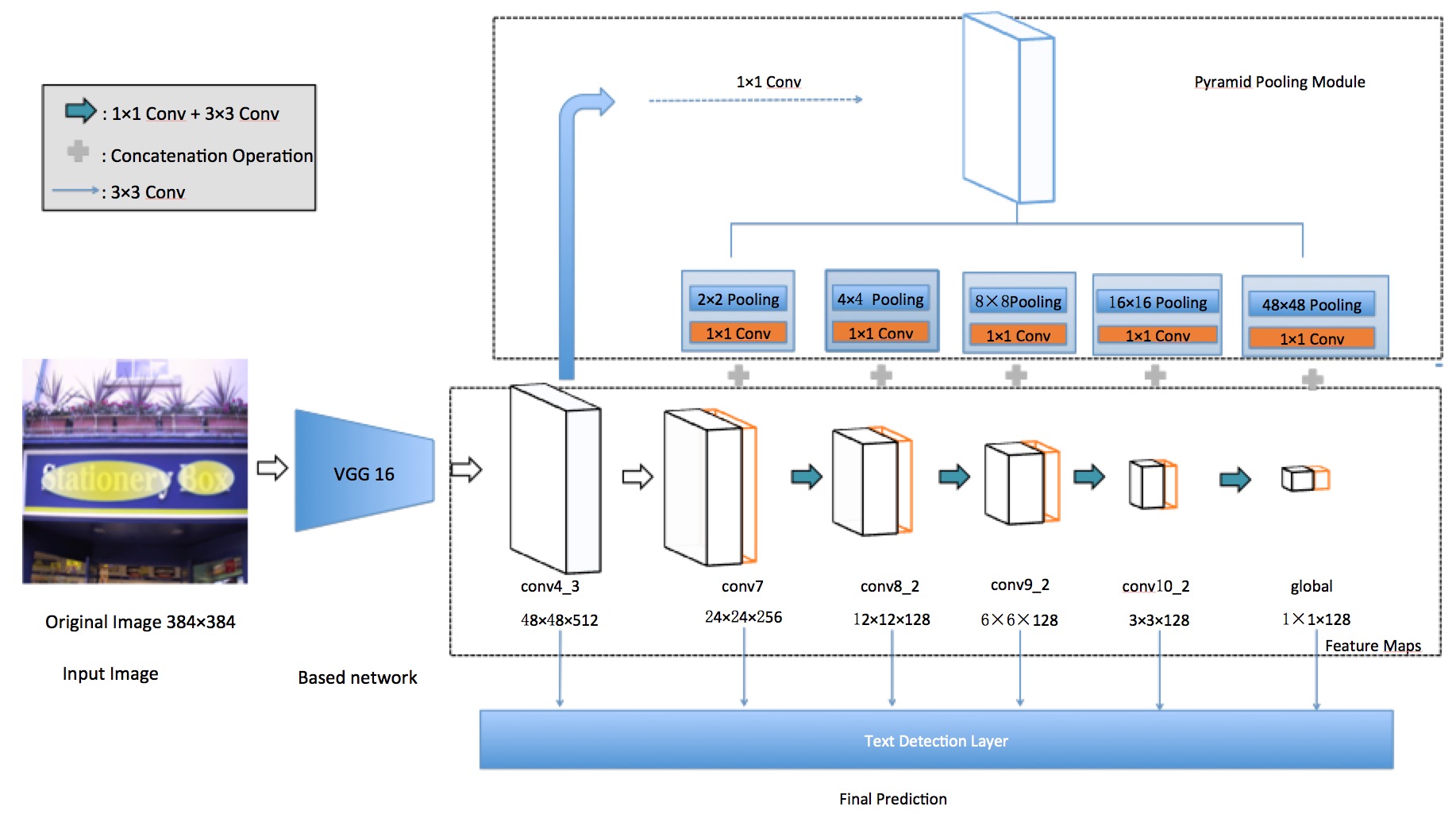}
\caption{\textbf{The Framework of the Proposed Method.} Given an input image with a size of 384$\times$384,VGG-16 base network outputs the first feature map from conv4\_3 layer. More feature maps with cascading sizes are extracted from extra layers following the first feature map. The first feature map is also used to produce different sub-region representations through the Pyramid Pooling Module. These representation layers are then concatenate feature maps with same size to output the final feature maps. Finally, those maps are fed into a convolution layer to get the final.}
\label{fig:1}
\end{figure*}

\indent Despite their promising performance in generic object detection, these methods suffered from bridging gaps between the data distributions of texts and generic objects. To enhance the generative abilities of existing deep models, \cite{gupta_synthetic_2016} proposed to naturally blend rendered words onto wild images for training data augmentation. The trained model based on such training data is robust to noises and uncontrolled variations. \cite{zhong_deeptext:_2016} and \cite{liao_textboxes:_2016} attempted to integrate region-proposal layers into the deep neural networks, which can generate text-specific proposals (e.g. bounding boxes with larger aspect ratios). In \cite{ma_arbitrary-oriented_2017}, the bounding-box rotation proposals were introduced to make the proposed model more adaptive for unknown orientations of texts in natural-scene images. Nevertheless, the aforementioned scene-text detectors either needed to consider a lot of proposal hypotheses, thus dramatically decreasing the computational efficiency, or utilize insufficient preset bounding-box characteristics to handle severe visual variations of scene-texts in unconstrained natural images.

\indent To address the aforementioned drawbacks of existing methods, we propose a novel proposal-free model for arbitrary-oriented text detection in natural images based on the circle anchors and the Single Shot Detector (SDD) framework. More specifically, we adopt circle anchors to represent the bounding boxes, which are more robust to orientation, aspect ratio, and scale variations, compared to the conventional rectangular ones. The Single Shot Detector (SSD), one of the state-of-art object detectors, is employed, considering its fast detection speed and promising accuracy in generic object detection. Besides the feature maps generated by the original SSD, we additionally incorporate a pyramid pooling module, which can build multiple feature representations on different spatial scales. By merging those different kinds of feature maps, both the local and global information can be preserved, such that texts in unconstrained natural scenes can be more reliably detected. Subsequently, the merged feature maps are fed into a text detection module, consisting of several fully connected convolutional layers, to predict confidential circle anchors. Furthermore, in order to overcome the difficulty of deciding positive points caused by unfixed sizes of circle anchors, we introduce a novel mask loss function by assigning those ambiguous points to a new class.
To obtained the final detection results, the Locality-Aware Non-Maximum-Suppression (LANMS) scheme  \cite{zhou_east:_2017} is employed. It should be noted that we do not utilize any proposal, which makes the proposed method more computationally efficient. 

\indent In summary, the contributions of our work mainly lie in three-fold:
\begin{itemize}[label={$\bullet$}]
    \item We propose a novel proposal-free method for detecting arbitrary-oriented texts in unconstrained natural scene images, based on the circular anchor representation and the Single Shot Detector framework. The circular anchors are more robust to different aspect ratio, scale, and orientation variations, compared to conventional rectangular ones. 
    \item We incorporate a pyramid pooling module into SSD, which can explore both the local and global visual information for robust text detection.
    \item We develop a new mask loss function to overcome the difficulty of deciding positive points caused by unfixed sizes of circle anchors, which can therefore improve the final detection accuracy. \end{itemize}

\section{Related Works}
  Character-based detection methods have already achieved state-of-art results on horizontal texts in relatively controlled and stable environments. Methods like those proposed in \cite{huang_robust_2014}, \cite{jaderberg_deep_2014}, \cite{neumann_real-time_2016}, and \cite{wang_end--end_2012} either detect individual characters by classification of sliding windows or utilize some form of connected-component and region-based framework such as the Maximally Stable Extremal Regions(MSER) detector.
  
\indent However, some of these methods might not be ideal for detecting scene-texts or multi-oriented texts as more and more environmental variations and uncertainties in terms of text distortion, orientation, occlusion, and noise are introduced. Detecting individual characters in close clusters or ones that blend into the background can also be challenging. Hence, many researchers decide to tackle the problem by approaching the task of text detection as object detection: treating words and/or text-lines as the target object.

\indent Region-based Convolutional Neural Network(R-CNN)\cite{ren_faster_2015}, Single Shot Detector(SSD)\cite{liu_ssd:_2016}, and segmentation-based Fully Convolutional Network(FCN)\cite{shelhamer_fully_2016} are frequently re-purposed for text detection because their superior speed and accuracy are better suited for the time and resource-constraining nature of the task. Expectedly, a majority of the cutting-edge research in scene-text detection, including ours, are based on one of the aforementioned object detection models, which we will analyze below.

\indent \textbf{Segmentation-based Methods:} Both \cite{zhang_multi-oriented_2016} and \cite{yao_scene_2016} accomplish semantic segmentation of text lines by utilizing Fully Convolutional Network(FCN), which has achieved great performance in pixel-level classification tasks. In \cite{zhang_multi-oriented_2016}, for example, pixel-wise text/non-text salient map is first produced via the FCN and subsequently, geometric and character processing is implemented to generate and filter text-line hypothesis. Although these methods can achieve state-of-art results even with scene-text detection in the wild, the requirement for a sophisticated post-processing step of word partitioning and false positive removal can be too time consuming and computationally intensive for real-world applications.

\indent A more recent method proposed in \cite{zhou_east:_2017}, however, seeks to make dramatic improvement in efficiency over \cite{zhang_multi-oriented_2016} and \cite{yao_scene_2016} by eliminating intermediate steps such as candidate aggregation and word partitioning in the neural network. Nevertheless, the inherent nature of the segmentation approach - dense per-pixel processing and prediction - is still a bottleneck that prevents segmentation-based methods from outperforming its competitors.

\indent \textbf{Region-Proposal-based Methods:} Although region-proposal based models like R-CNN have already been a state-of-art object detector, it can not be implemented for the purpose of text detection without modifications since its anchor box design is not ideal for the large aspect ratio of words/text-lines. \cite{zhong_deeptext:_2016} addresses this problem by proposing a novel Region Proposal Network(RPN) called Inception-RPN, which contains a preset of text characteristic prior bounding boxes to generate text-specific proposals and thus filtering out low-quality word regions. 

\indent However, \cite{zhong_deeptext:_2016} only performs well on horizontal texts since bounding box characteristics are extremely unpredictable for scene-texts in the wild; multi-oriented and distorted texts can create countless possibilities and variations of bounding-box size, shape, and orientation.

\indent To address this challenge, some researchers designed novel region-proposal methods: the rotation proposal method in \cite{ma_arbitrary-oriented_2017} has the ability to predict the orientation of a text line and thus generate inclined bounding-boxes for oriented texts, while the quadrilateral sliding windows in \cite{liu_deep_2017} create a much tighter bounding-box fit around text regions, thus dramatically reduce background noise and interference. On the other hand, some researchers propose methods to modify model architecture like the one proposed in \cite{dai_deformable_2017}, which adds 2D offsets in the standard convolution to enable free form deformation of the sampling grid, and the one proposed in \cite{he_deep_2017}, which utilizes direct bounding-box regression originating from a center anchor point in a proposal region.

\indent \textbf{SSD-based Methods.} SSD-based method is highly stable and efficient in generating word proposals because SSD is one of the fastest object detector that is also as accurate as slower region-proposal based models like R-CNN. However, SSD possesses similar shortcomings in terms of anchor box design when it comes to scene-text detection. Thus, \cite{liao_textboxes:_2016} supplements SSD with "textbox layers" that can generate bounding-boxes with larger aspect ratios and simultaneously predict text presence and bounding boxes. Unfortunately, this method only works on horizontal texts, and not scene-texts. \\
\indent Thus in this paper, we attempt to solve the aforementioned limitations of previous detection models by utilizing a proposal-free method based on circular anchors and the SSD framework. Our method is computationally more efficient than both segmentation-based and Region-Proposal-based models because the removal of the region-proposal layer in our network. Our method also improves upon the existing SSD-based method by having the ability to detect both arbitrary-oriented texts and generic objects.

\section{Method}
In this section, we will describe the details of our proposed model - ArbiText. We will first introduce the framework and network architecture of our method. Subsequently, we will elaborate on the key components such as the circle anchor representation and the proposed loss function. 
\begin{figure}
\includegraphics[width=\linewidth]{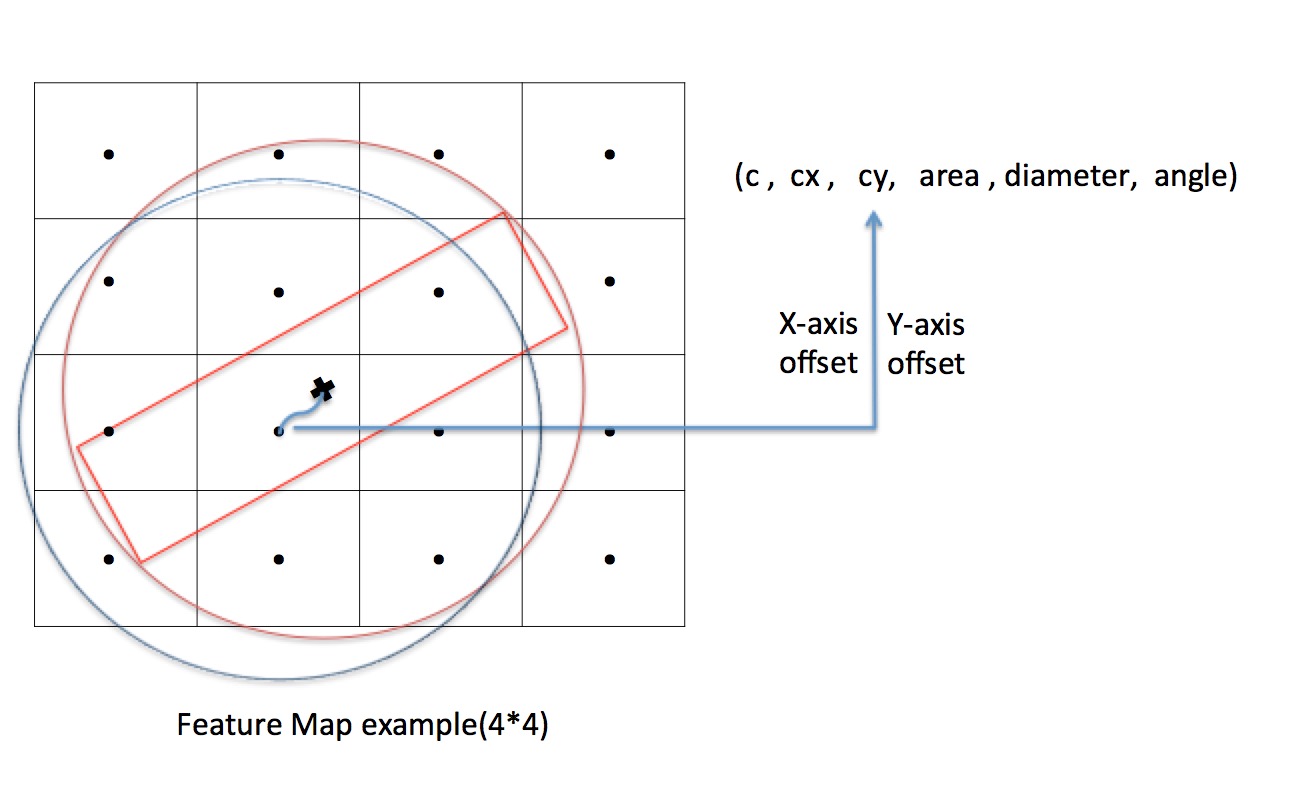}
\caption{\textbf{Matching of Circle Anchors} The red circle is generated from the ground-truth coordinates, and the blue circle anchor is the one with same size with red circle on the feature point. The blue circle which associates the red one by a vector(cx, cy, area, diameter, angle) where cx and cy are offset between centers of circles.}  
\label{fig:2}
\end{figure}

\subsection{Model Framework}Our proposed method, in essence, is a multi-scale, proposal-free framework based on the Single Shot Detector. As shown in Fig.2, our model mainly consists of the following four components: 1) the backbone-based network for converting original images into dense feature representations; 2) the feature maps component with cascading map size for detecting multi-scale texts; 3) the Pyramid Pooling Module \cite{zhao_pyramid_2016} for extracting sub-region feature representations; and 4) the final text detection layer for circle anchor prediction.

\indent We adopt VGG-16 \cite{simonyan_very_2014} as our base network, and utilize the 6 feature maps at the conv4\_3, conv7, con8\_2, conv9\_2, conv10\_2 and global layers. However, local information is lost as layer goes deeper and deeper, which results in poor detection precisions especially on texts with complex contextual information.

\indent Inspired by \cite{zhao_pyramid_2016}, we introduced the Pyramid Pooling Module to leverage low-level visual information even in deeper layers. This module fuses feature maps with different pyramid scales. As shown in Fig. 2, the first feature map from the based network is separated on pyramid levels into different sub-regions and output pooled representations after a $1*1$ convolution layer. Thus, the low-level information of the original image could be preserved in multi-scale level feature maps, which will be further concatenated with ones of the same size to form the final feature map for text detection. By merging these two types of features, both the local and global visual information can be explored.

\indent Finally, the text detection layer applies a $3*3$ convolution kernel on the fused feature map to output prediction on the text bounding box.

\subsection{Circle Anchors}

As illustrated in Fig. 3, instead of the traditional rectangular anchors, we use circle anchors to represent the bounding box. Specifically, a bounding box can be represented by a 5-dimensional vector ($x$, $y$, $a$, $r$, $\theta$), where $a$, $r$, and $\theta$ denotes the area, radius, and rotated angle of a circle anchor.

\begin{equation}
\begin{gathered}
\alpha = \frac{1}{2}arcsin(\frac{a}{(2r^2)})  \\
x_2 = r\cdot cos(\alpha + \theta), y_2 = r \cdot sin(\alpha + \theta) \\
x_3 = r \cdot cos(\alpha - \theta), y_3 = r \cdot sin(\alpha - \theta) \\
x_1, y_1 = -x_3, -y_3,x_4, y_4 = -x_2, -y_2 \\
x_1, x_2, x_3, x_4 = x_1 + x, x_2 + x, x_3 + x, x_4 +x \\
y_1, y_2, y_3, y_4 = y_1 + y, y_2 + y, y_3 + y, y_4 +y 
\end{gathered}
\end{equation}

On a feature map of size $(w * w)$, location, denoted $(i,j)$, associates a circle anchor $C_0$ with $(c, \Delta x, \Delta y, \Delta a, \Delta r, \Delta \theta)$, indicating that a unique circle anchor, represented by $(x, y, a, r, \theta)$, is detected with confidence $c$, where

\begin{equation}
\begin{gathered}
x = \Delta x \cdot \frac{r_a}{w} + j,y = \Delta y \cdot \frac{r_a}{w} + i \\
r = exp(\Delta r) \cdot \frac{r_a}{w} \\
a = exp(\Delta a) \cdot \frac{r_a}{w} \\ 
\theta = \Delta \theta 
\end{gathered}
\end{equation}

\indent Here, we use the area $a$ and radius $r$ for computational stability. Also, we multiply each value by a factor $\frac{r_a}{w}$ where $r_a$ =1.5. 

\indent The angle $\theta$ is the intersection angle between the long edge of the bounding box and the horizontal axis. Thus, the value of $\theta$ ranges from $\ang{-90}$ to $\ang{90}$.

\indent In a deep neural network, each layer has a receptive field that indicates how much contextual information we can utilize. Although the circle anchor representation is invariant to scale variations, \cite{zhou_object_2014} has shown that feature maps have limited receptive fields that are much smaller than theoretical ones, especially on high-level layers. As a result, if we do not utilize multi-scale feature maps, the detection scope of the proposed circle anchor representation will be restricted. And considering the size of the extra feature layers, this operation only adds a small amount of computational cost. 
\subsection{Training Labels Rebuilding and the Loss Function Formulation}.

\begin{figure}
\subcaptionbox{\label{sfig:testa}}{\includegraphics[width=4.0cm]{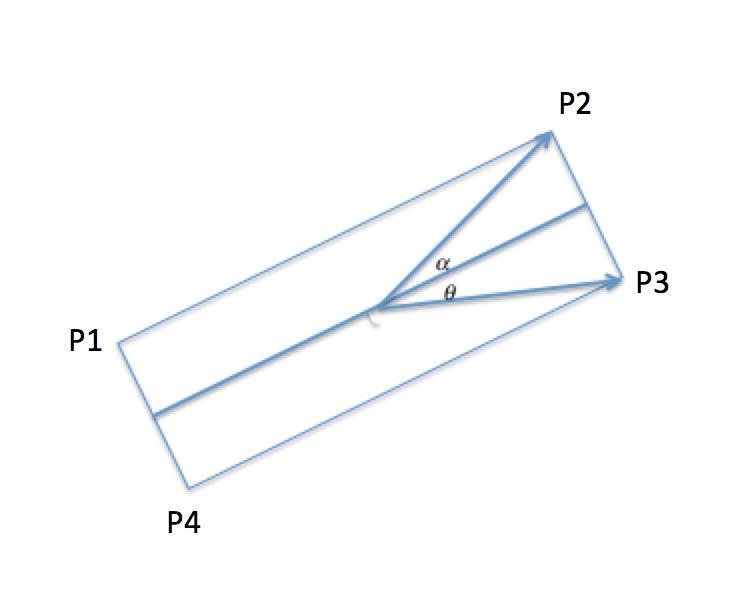}}
\subcaptionbox{\label{sfig:testb}}{\includegraphics[width=4.0cm]{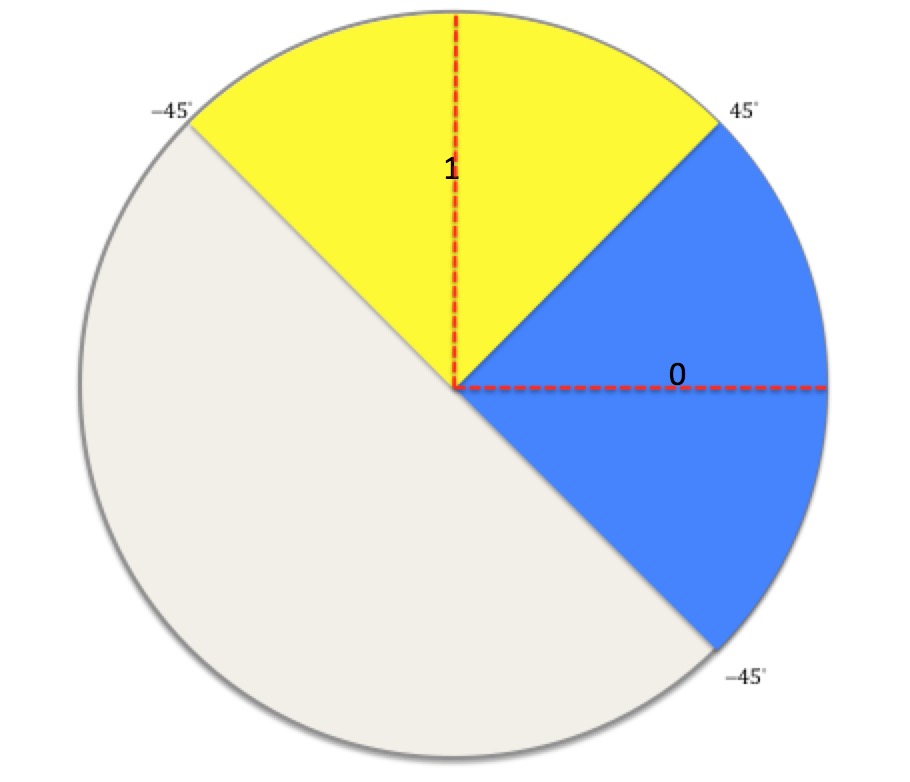}}
\caption{ (a) shows the the coordinates of a rectangle. Given the area and diagonal of rectangle, $p2$ and $p3$ can be calculated by rotation angles. $p1$ and $p4$ are diagonal points of $p3$ and $p1$, respectively, which also have negative values. (b) shows the predictable vertical flag, which is $0$ if the angles of the bounding box are between \ang{-45} and \ang{45}; otherwise, it is set to $1$.}
\end{figure}

\begin{figure}
\includegraphics[width=\linewidth]{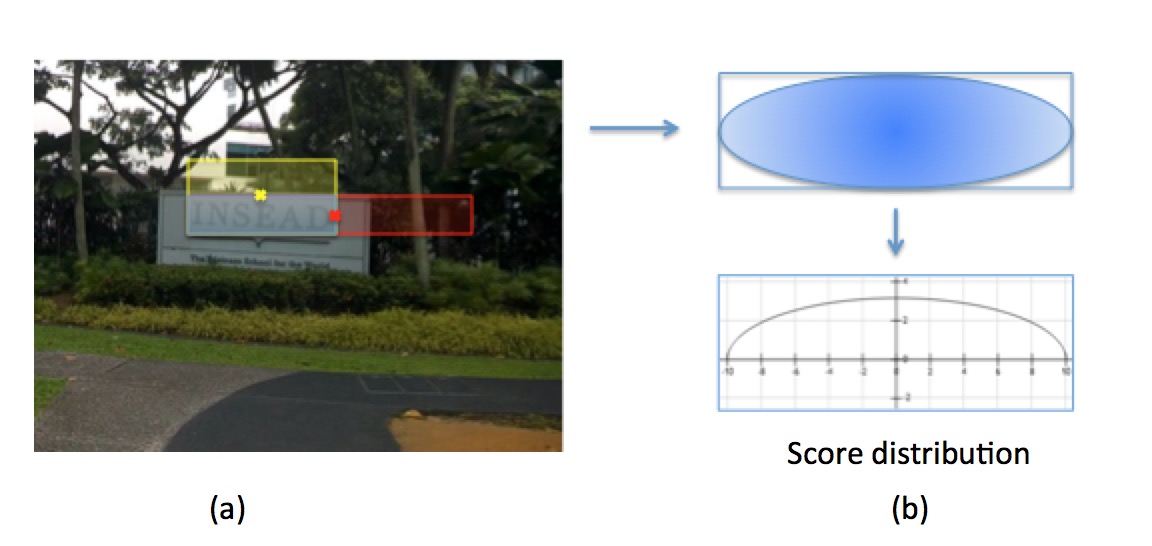}
\caption{\textbf{Score Distribution.} (a) shows a bounding box the text. The red and yellow rectangles are possible bounding boxes that have maximum IOU overlap scores with ground truth . (b) shows the eclipse-shape like score function we use in \textbf{ArbiText}.}
\end{figure}

\begin{figure}
\includegraphics[width=\linewidth]{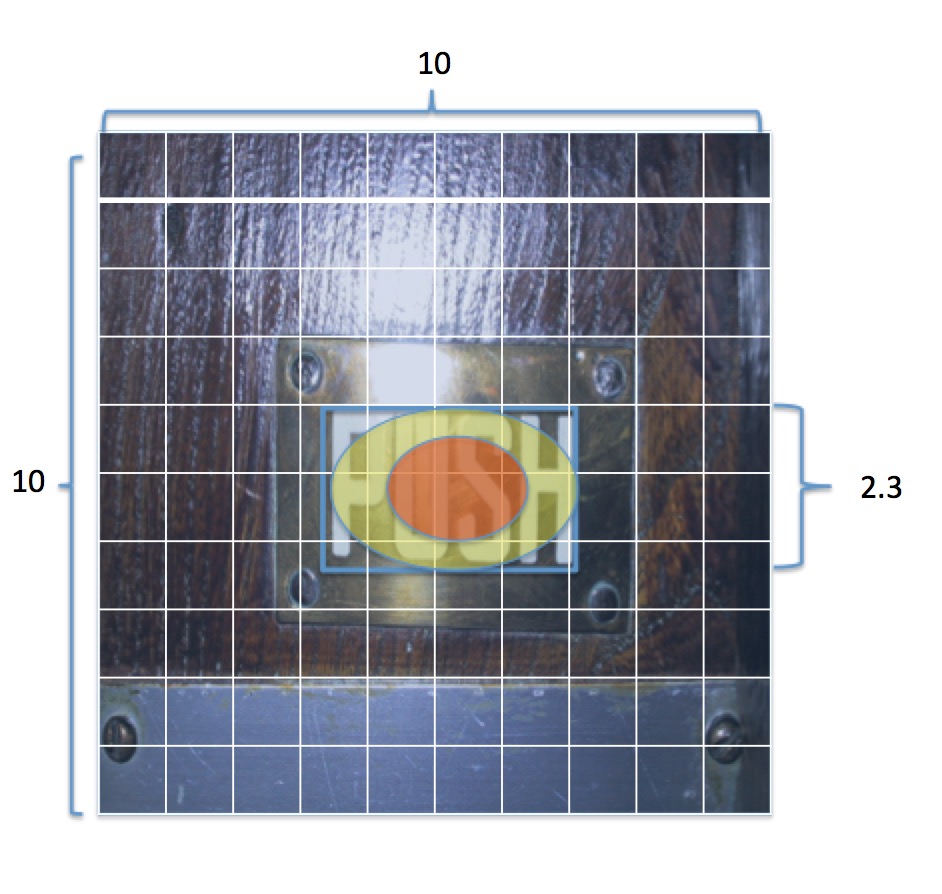}
\caption{\textbf{The Selection of Ground Truth.} The blue rectangle is the bounding box and the $10*10$ grid represents a feature map with the same size. Only the feature points in the red eclipse are labeled as positives; the points outside of the yellow eclipse are labeled as negatives. The feature points in the yellow region will be labeled as their own separate class.}
\end{figure}
For a SSD-based method, all points on a feature map will potentially be used for minimizing a specific loss function. Each feature point needs to be labeled as either “positive” or “negative”. Specifically, in SSD, the points that are labeled as “positive” are chosen from the regions where the overlap between the default anchor and ground-truth bounding box is larger than 0.5. However, there is no default anchor in our method, but we can still calculate a confidence score for each point. As illustrated in Fig. 5 (a), the feature point on the edge of the bounding box can have a maximum overlap of 0.5(bounding boxes are colored in red and yellow). So the score follows an eclipse distribution (as illustrated in Fig. 5 (b)). The scores at the center of the eclipse have a maximum value of 1.0 and it decreases to 0.5 when the points reach the edge. We use a semi-ellipse as the function to compute the score for each point. As a result, the points outside of the eclipse have a score value 0. \\
\indent Imagine an eclipse score function which has rotation angle $\theta$, the semi-major axes and semi-minor axes have length of $\frac{w}{2}$ and $\frac{h}{2}$, respectively, where $w$ and $h$ are the width and height of the bounding box. Thus, the score function can be represented as:
\begin{equation}
\begin{gathered}
A \cdot (x)^2 + B\cdot(x)\cdot(y) + C\cdot(y)^2 + F\cdot s^2 = F \\
a = \frac{w}{2} ; b = \frac{h}{2} \\
A = a^2\cdot sin^2(\theta) + b^2\cdot cos^2(\theta) \\
B = -2(a^2-b^2) \cdot sin(\theta)\cdot cos(\theta) \\
C = a^2\cdot cos^2(\theta) + b^2\cdot sin^2(\theta) \\
F = a^2\cdot b^2
\end{gathered}
\end{equation}

where $x$,$y$ are the distances between a feature point and the center of the bounding box respectively,s is score. According to the score function, all points inside the eclipse have a score greater than 0.5. However, the closer the point is to the edge of the bounding box, the large the noise outside of the bounding box will be, which could make the training of networks harder to converge. Thus, only the points with score large than a threshold $\alpha$ will be treated as positives (as shown in Fig. 6, only points inside the red zone are labeled as "positive"). Points outside of the bounding box will be labeled as "negative". For those points with a score between 0.5 and $\alpha$, we assign them an additional label. Thus, there will be a total of $(N+1)$ classes(without background). This additional class will only be involved in calculating the classification loss.\\
\indent The feature maps with different sizes can detect texts with different scales. A default box will be labeled as positive if 
\begin{equation}
1<\frac{h_g}{c_h} < 4.
\end{equation}
where $h_g$ is the height of bounding box and $c_h$ is the height of cell on feature map.
\indent  For training, we use the following objective loss function:
\begin{equation}
\begin{split}
L(mask, c, l, g) = \frac{1}{N_{cls}}\sum_{i=0}^{N+1}mask_{i}\cdot L_{cls}(c_i) + \\ \lambda_1\frac{1}{N_{reg}}\sum_{i=1}^{N}mask_{i}\cdot L_{loc}(l,g) + \\
\lambda_2\frac{1}{N_{reg}}\sum_{i=1}^{N}mask_{i}\cdot L_{cls}(vertical)
\end{split}
\end{equation}
where $N$ is the number of object categories, $l$ is the prediction location, and $g$ is the ground-truth location. For each class, $mask_i = 1$ if the corresponding point is labeled as "positive" and belongs to the $i$-th class. is the loss for vertical bounding box classification that only includes the positive points. We adopt L1 loss for smoothing and Softmax loss as the classification losses.

\begin{figure*}[t]
\begin{multicols}{3}
    \includegraphics[width=\linewidth]{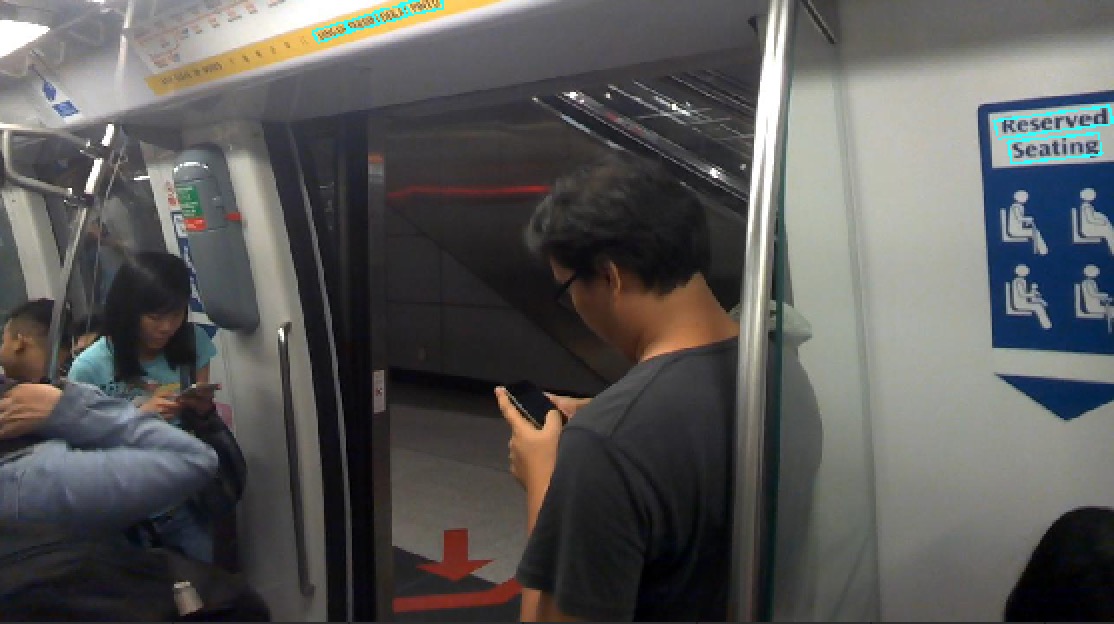}\par 
    \includegraphics[width=\linewidth]{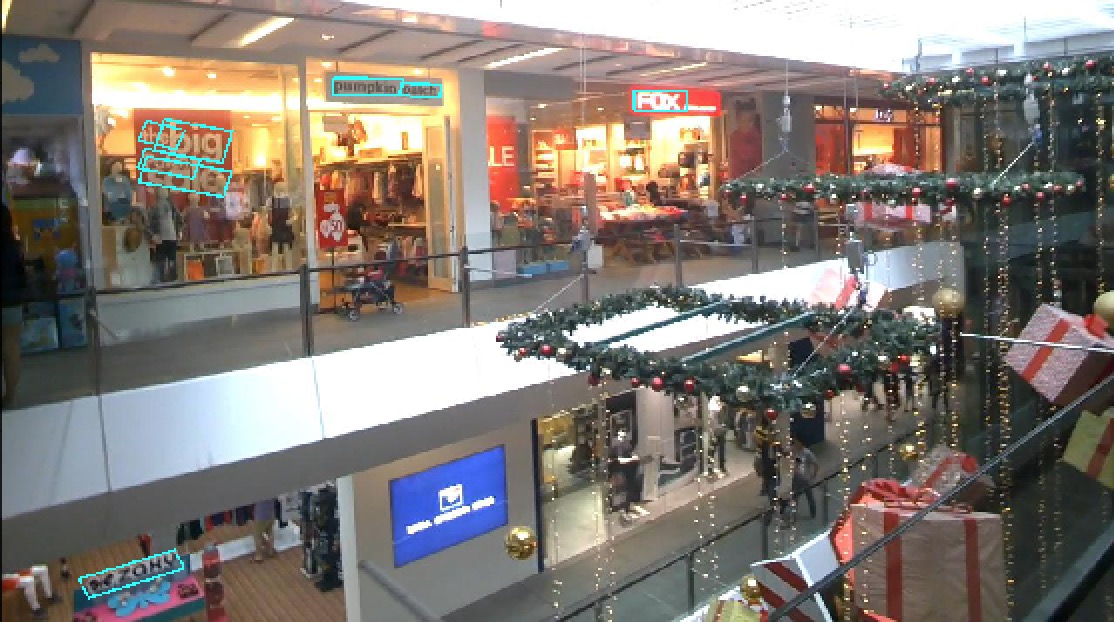}\par
    \includegraphics[width=\linewidth]{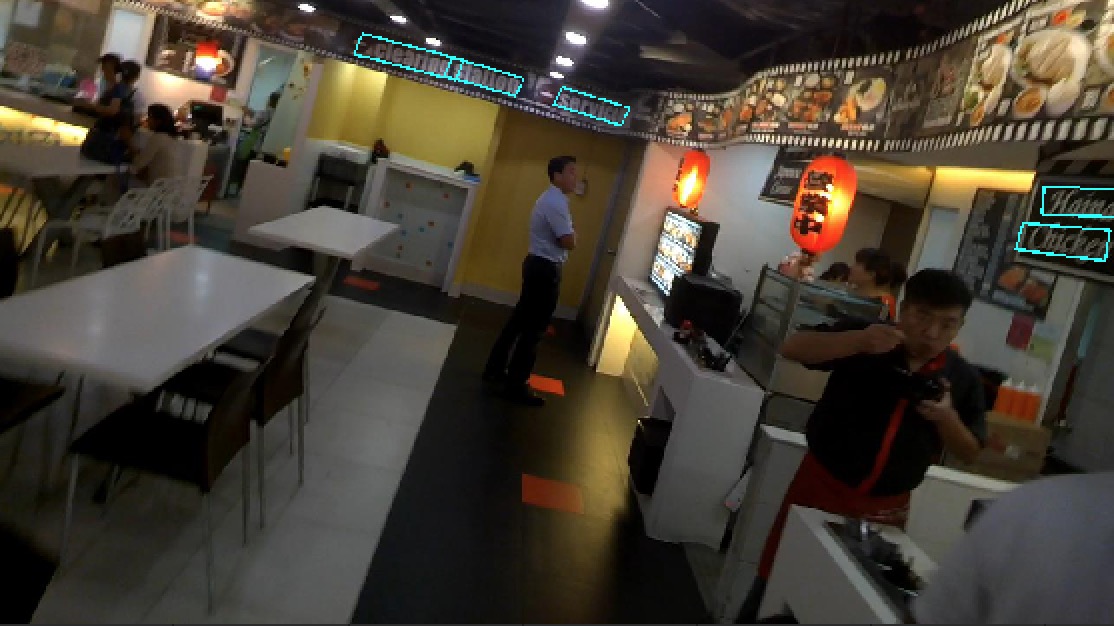}\par
\end{multicols}
\begin{multicols}{3}
    \includegraphics[width=\linewidth]{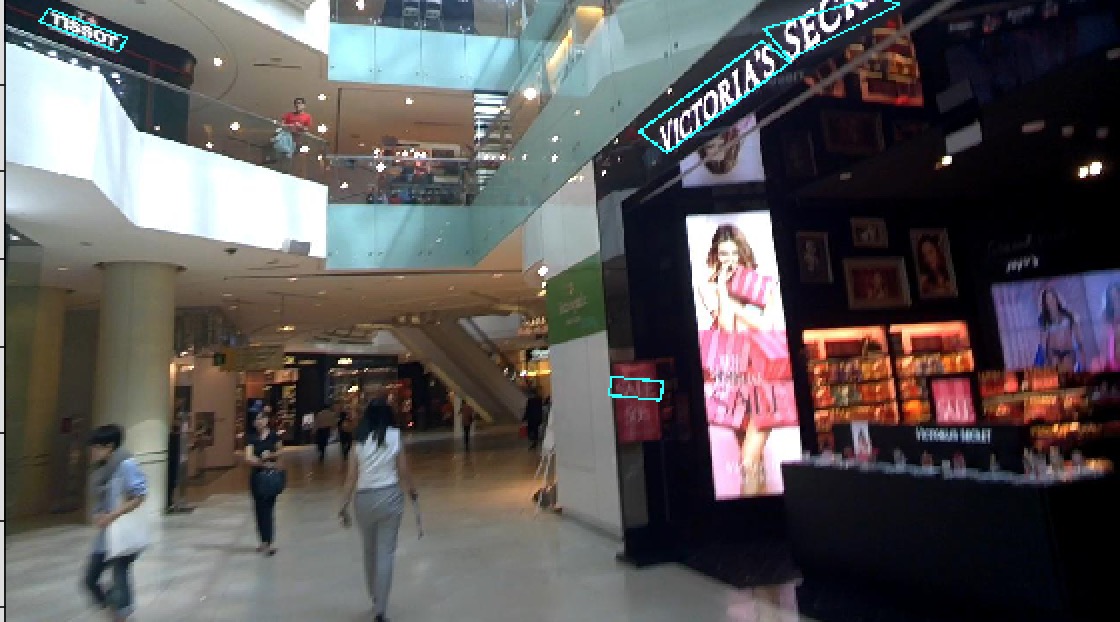}\par
    \includegraphics[width=\linewidth]{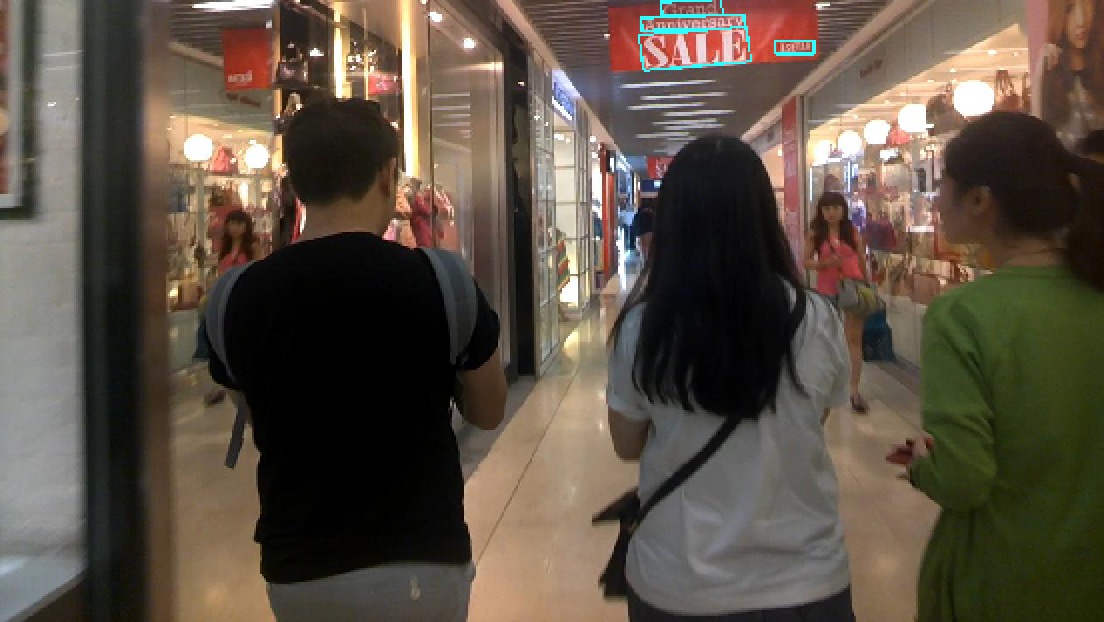}\par
    \includegraphics[width=\linewidth]{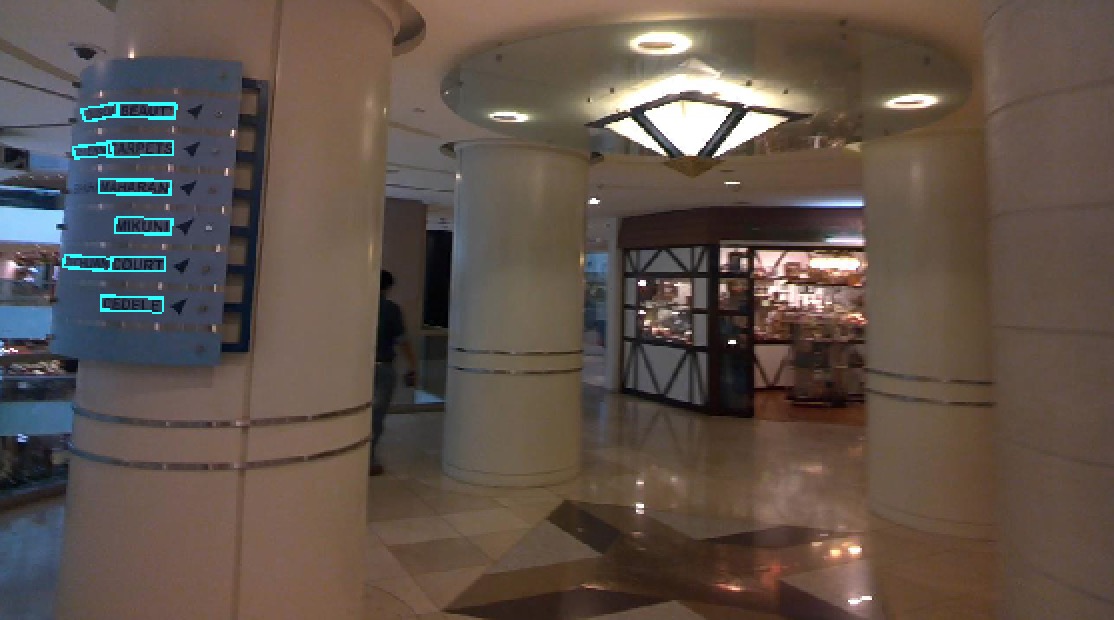}\par
\end{multicols}
\begin{multicols}{3}
    \includegraphics[width=\linewidth]{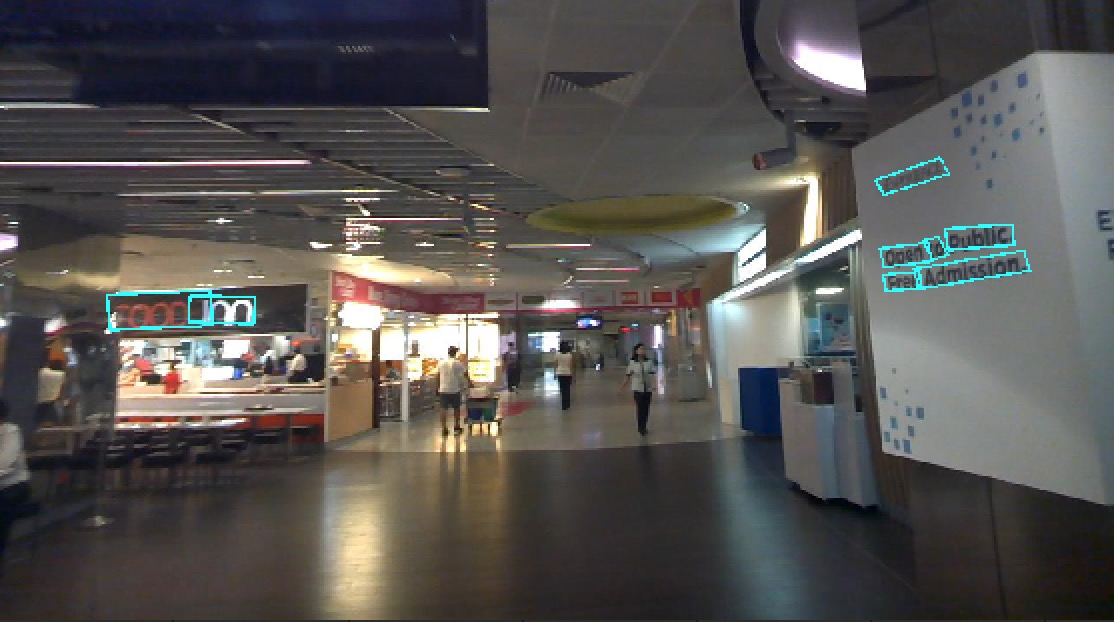}\par
    \includegraphics[width=\linewidth]{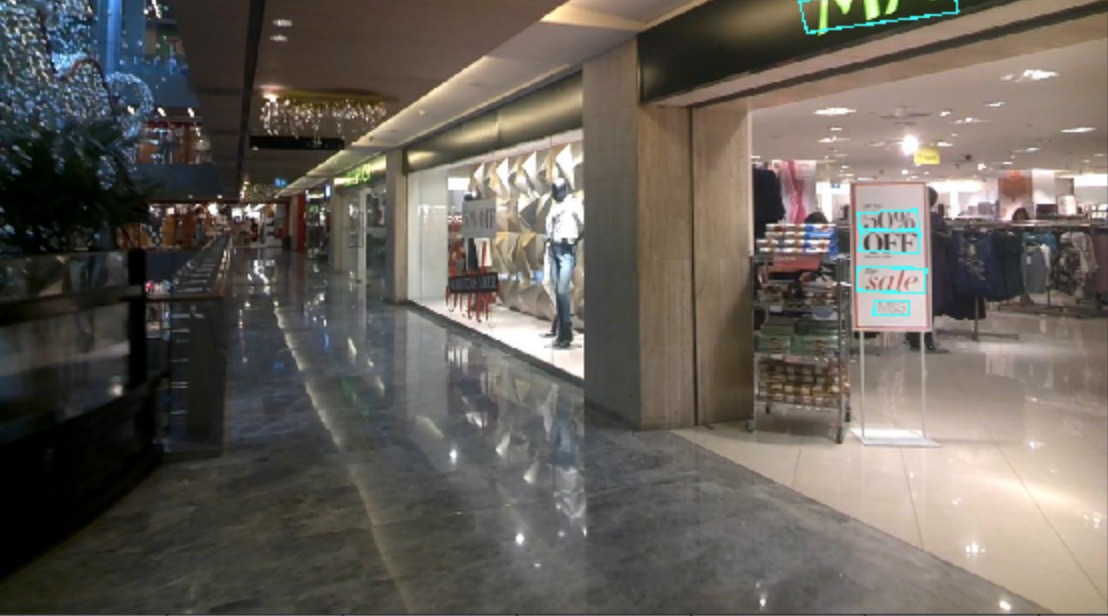}\par
    \includegraphics[width=\linewidth]{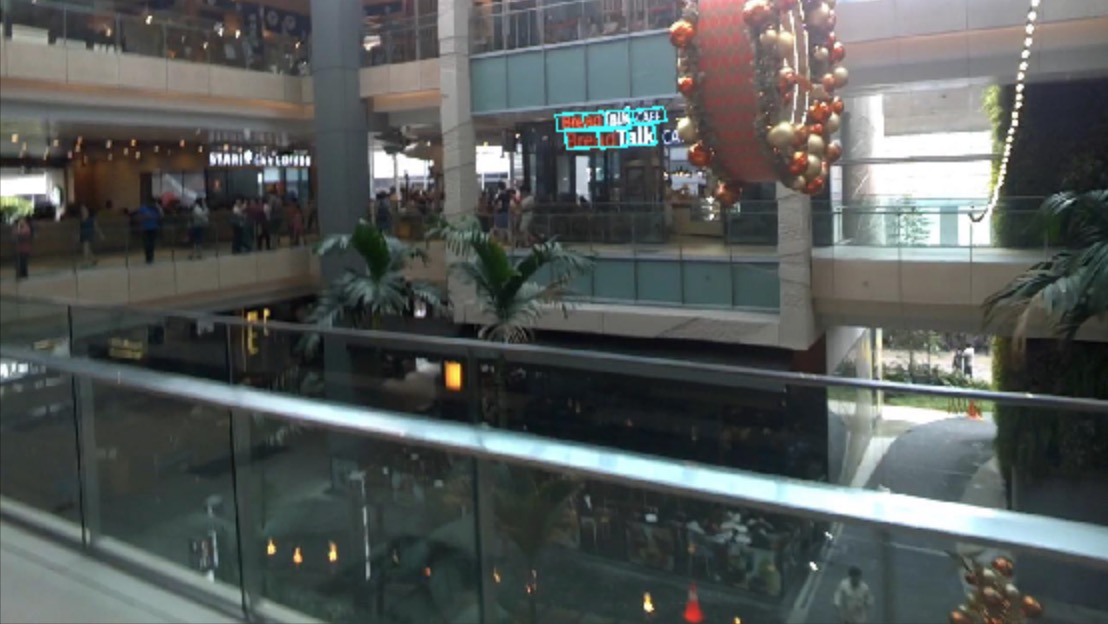}\par
\end{multicols}
\caption{\textbf{Results of \textbf{ArtTex} on the ICDAR2015 dataset.} Blue rectangles are detected text regions by using \textbf{ArtTex}. }
\end{figure*}

\section{Experiments}

\subsection{Datasets}
In order to evaluate the performance of the proposed method, we ran experiments on two benchmark datasets: the ICDAR 2015 dataset and the MSRA-TD500(TD500) dataset. 

\textbf{SynthText in the Wild}\cite{gupta_synthetic_2016} 
dataset contains more than 800,000 synthetic images created by blending rendered words on wild images. Only samples with width larger than $10$ pixels are chosen for training.

\textbf{ICDAR 2015}\cite{zhou_icdar_2015} incidental text dataset is from Challenge 4 of ICDAR 2015 Robust Reading Competition that includes 1000 training images and 500 testing images. Since those images are collected by Google Glasses, they suffer from motion blur. The blurry texts have a label of "\#\#\#" and are excluded from our experiment. We also included training and testing images from the ICDAR 2013 dataset\cite{karatzas_icdar_2013}, which helps us in building a more robust text detector.

\textbf{MSRA-TD500(TD500)}\cite{cong_msra_2012} is a multilingual dataset that includes oriented texts in both Chinese and English. Unlike ICDAR 2015, texts in MSRA-TD500 are annotated at the text-line level and the images were captured more formally, thus texts are much clearer and standardized. There are a total of 500 images, 300 of them were used as training data and 200 were used as testing data.

\subsection{Implementation Details}

\textbf{Base network}
In our experiment, we uses a pre-trained VGG-16 as our based network. This network is widely used in object detection tasks. All images are resized to $384*384$ after data augmentation. We extracted five layers with cascading resolution as our feature maps, which are conv4\_3, conv7, conv8, conv9, and conv10. We first trained our model on the SynthText dataset for 50,000 iterations with a learning rate of 0.001. Then, we fine-tuned our model using the other datasets with a 0.0005 learning rate. The details of training different datasets are described in later sections. We tested our model on different $\alpha$ values and we discovered that our model achieves optimal performance when $\alpha$ is set to $0.7$ for text detection. 

\textbf{Locality-Aware NMS}
In the post-processing stage, bounding boxes with a confidence score greater than 0.5 will be used to produce the final output by NMS merging. The naive NMS has $\mathcal{O}(n^2)$ computational complexity, which is not ideal for real-world applications. We adopt Locality—Aware NMS  \cite{zhou_east:_2017} to improve speed of merging bounding boxes.

\textbf{Hard Negative Mining}
Hard Negative Mining is essential for SSD-based methods because of the imbalance between positive and negative training samples. We adopt the same configuration in SSD\cite{liu_ssd:_2016} by selecting the top $3K$ negative training samples, where $k$ is number of positive training samples. Thus, we adopt Locality-Aware NMS\cite{zhou_east:_2017} in our experiment. This algorithm can produce bounding boxes with greater precision in shorter time.

\textbf{Data Augmentation}
    We utilize a data augmentation pipeline that is similar to the one in SSD to make our model more robust against different text variations. The original image is randomly cropped into patches. The crop size is chosen from [0.1, 1] of original image size. Each sample patch will be horizontally flipped with a probability of $0.5$. In order to balance samples of different orientations, we also augmented datasets by randomly rotating images by $\theta$ degrees. $\theta$ is randomly chosen from the following angle set: (-90, -75, -60, -45, -30, -15, 0, 15, 30, 45, 60, 75, 90). 
\begin{figure*}[t]
\begin{multicols}{3}
    \includegraphics[width=\linewidth]{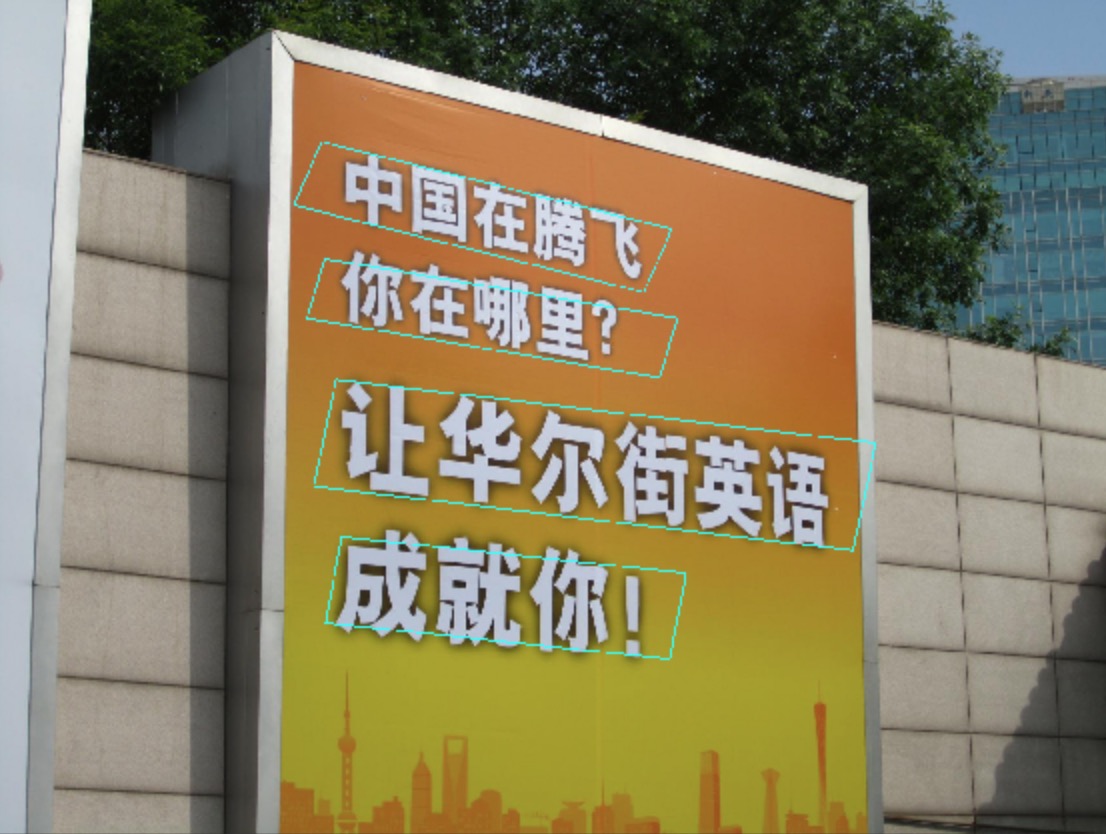}\par 
    \includegraphics[width=\linewidth]{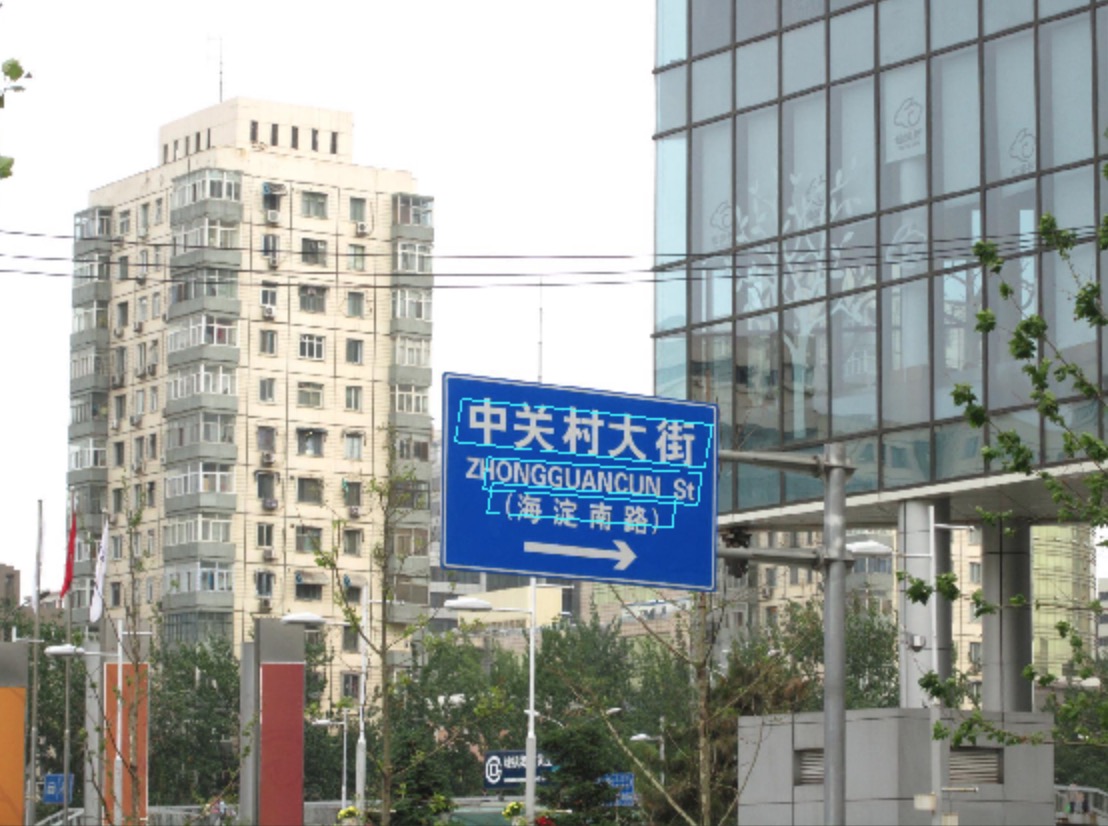}\par
    \includegraphics[width=\linewidth]{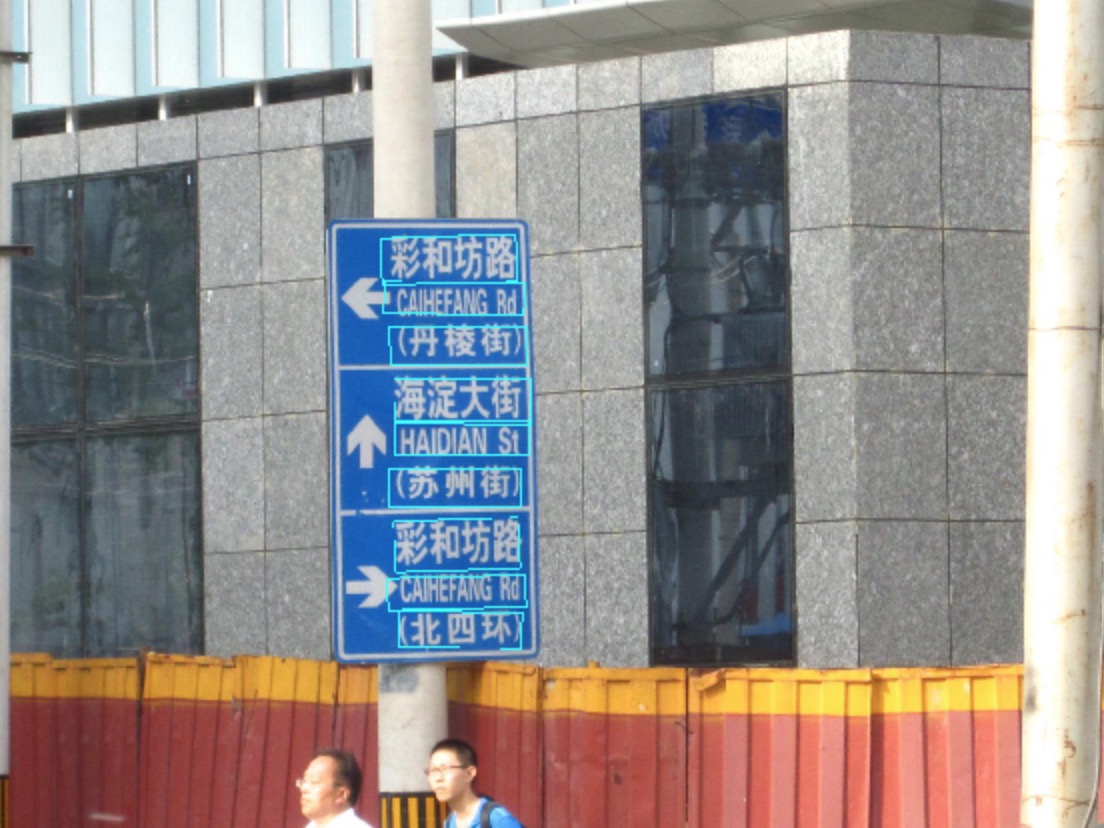}\par
\end{multicols}
\begin{multicols}{3}
    \includegraphics[width=\linewidth]{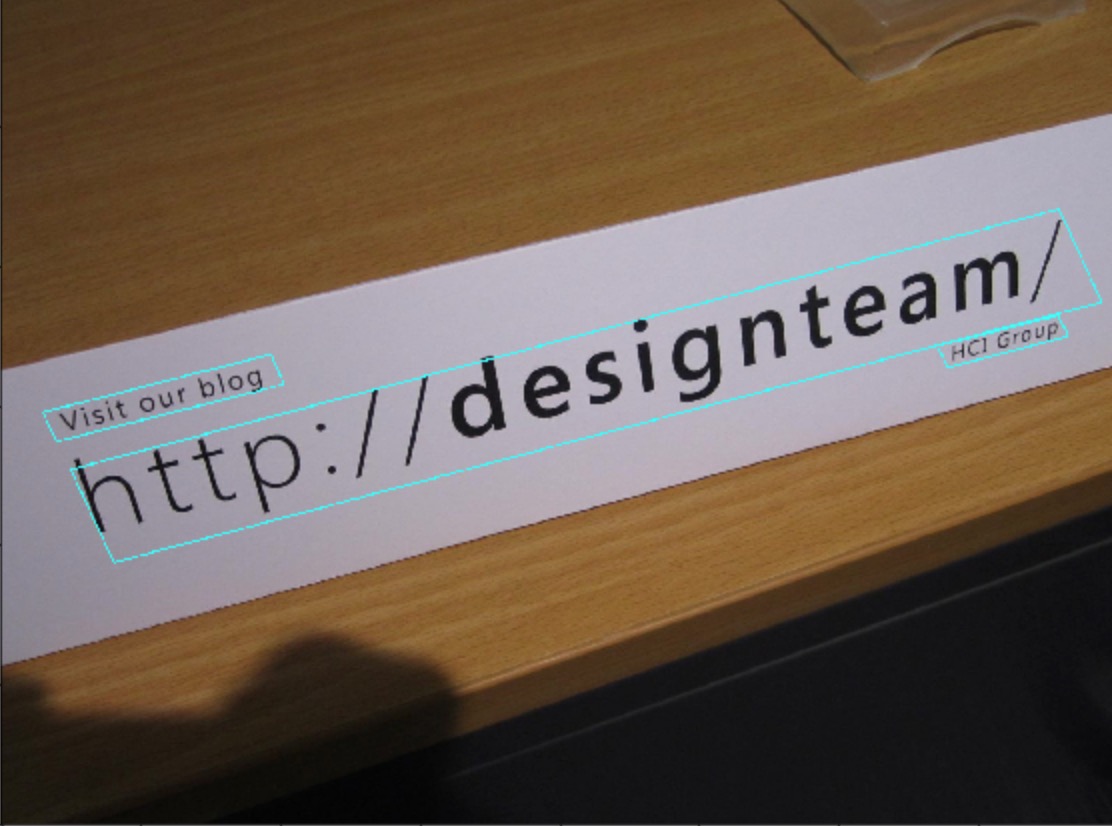}\par
    \includegraphics[width=\linewidth]{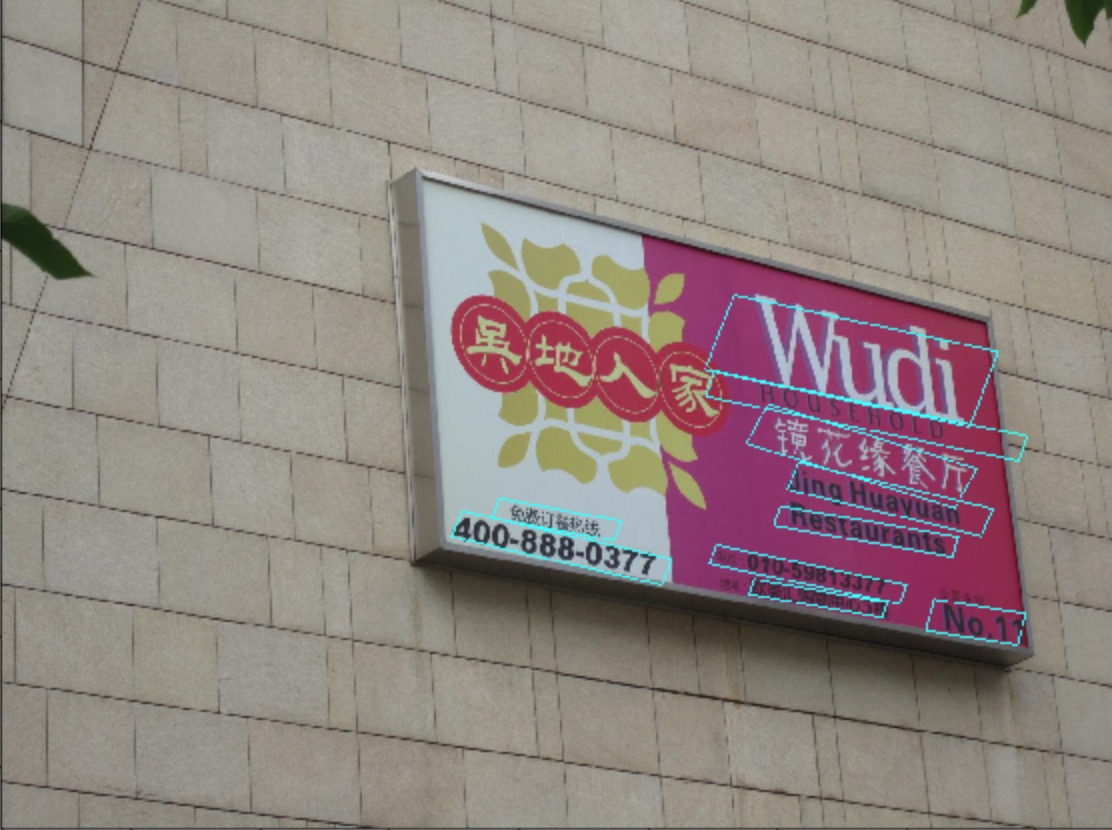}\par
    \includegraphics[width=\linewidth]{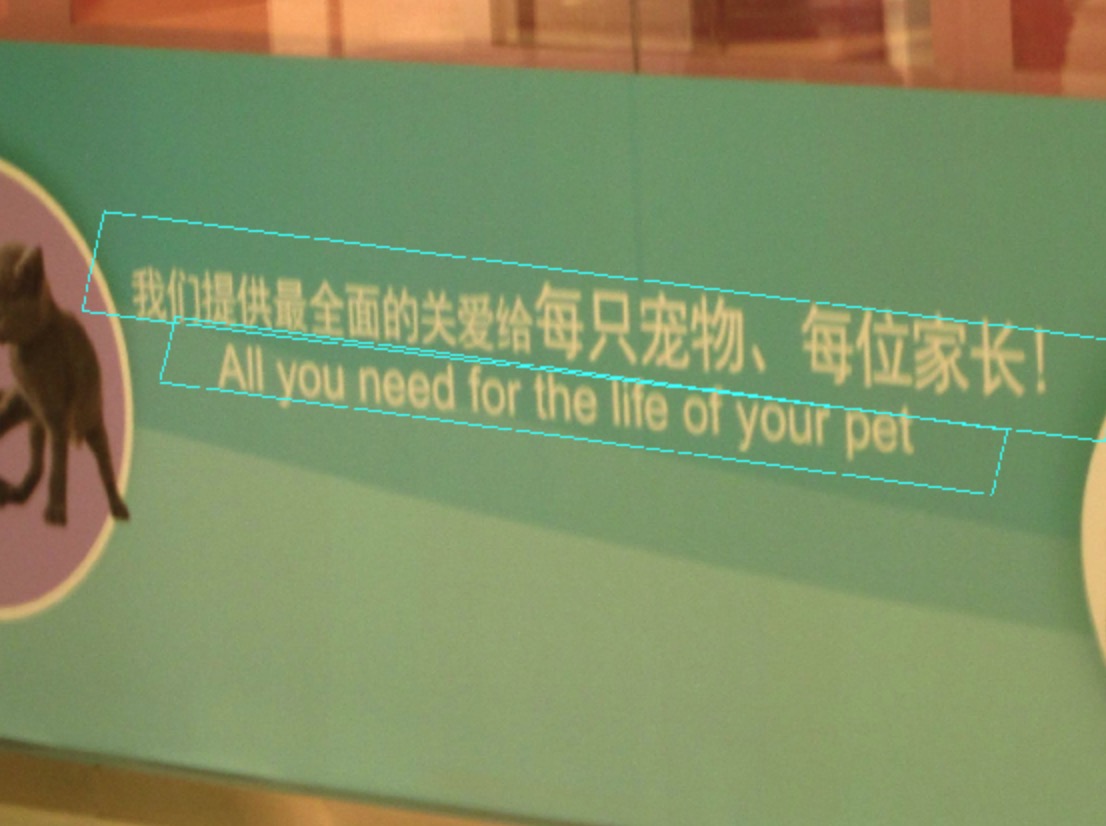}\par
\end{multicols}
\caption{\textbf{Results of \textbf{ArtTex} on the MSRA-TD500 dataset.} Blue rectangles are detected text regions by using \textbf{ArtTex}. }
\end{figure*}

\begin{table}
    \caption{Comparison results of various methods on the \textbf{ICDAR 2015 Incidental Text} dataset }
    \begin{center}
    \begin{tabular}{|l|c|c|c|}
    \hline    \textbf{Method}&\textbf{Precision}&\textbf{Recall}&\textbf{F-score}\\
    \hline\hline
    HUST\_MCLAB & 47.5 & 34.8 & 40.2 \\\hline
    NJU\_Text & 72.7 & 35.8 & 48.0 \\\hline
    StradVision-2 & 77.5 & 36.7& 49.8 \\\hline
    MCLAB\_FCN\cite{zhang_multi-oriented_2016} &70.8&43.0&53.6 \\\hline
    CTPN\cite{tian_detecting_2016} &51.6&74.2&60.9 \\\hline
    Megcii-Image++&72.4&57.0&63.8 \\\hline
    Yao \textit{et al.}\cite{yao_scene_2016}&72.3&58.7&64.8 \\\hline
    Seglink \cite{shi_detecting_2017} & 73.1 & \textbf{76.8} & 75.0 \\\hline
    ArbiText  & \textbf{79.2} & 73.5 & \textbf{75.9} \\\hline
    \end{tabular}
    \end{center}
\end{table}

\subsection{Detection Results}
\subsubsection{Detecting Oriented English Text}

First, our model is tested on ICDAR 2015 dataset. The pre-trained model is fine-tuned using both the ICDAR 2013 and the ICDAR 2015 training datasets after 20k iterations. Considering all images in ICDAR 2015 have high resolution, testing images are first resized to $768*768$. The threshold $\alpha$ is set to 0.7, similar to the one used in the pre-training stage. Performance is evaluated using the official off-line evaluation scripts. \\
\indent We list the results of our model along with other state-of-art object and text detection methods. The results were obtained from the original papers. The best result of this dataset was obtained by Seglink\cite{shi_detecting_2017}, which achieved a F-measure score of 75.0\%. However, our model obtained a score of 75.9\%. The improvement comes from the high precision rate we obtained, which outperforms the second highest model by 6.1\% \\
\indent Figure.7 shows several detection results taken from the testing dataset of ICDAR 2015. Our proposed method ArbiText can distinguish and localize all kinds of scene text in noisy backgrounds. \\

\subsubsection{Detecting Multi-Lingual Text in Long Lines}
We further tested our method on the TD500 dataset consists of long text in English and non-Latin scripts. We augmented this dataset by doing the following: 1) Randomly place an image on a canvas of $n$ times of the original image size filled by mean values where $n$ ranges from 1 to 3. 2)We applies random crop according to the overlap strategy described in section 4.3. Thus, we obtained enough images for training. The pre-trained model is fine-tuned for 20K iterations. All images are resized to $384*384$, which is consistent with the training stage. The experiment has demonstrated that this technique can dramatically increase detection speed without losing much precision. As illustrated in Table 2, ArbiText achieved comparable F-measure scores with other state-of-the-art methods. However, benefiting from lighter network architecture and simplified anchors mechanism, ArbiText has the highest FPS of 12.1. \\
\indent Figure.8 shows ArbiText can detect long lines of text in mixed languages(English and Chinese) without changing any parameters or structures. 

\begin{table}
    \caption{Comparison results of various methods on the \textbf{MSRA-TD500} dataset}
    \begin{center}
    \begin{tabular}{|l|c|c|c|c|}
    \hline
    \textbf{Method}&\textbf{Precision}&\textbf{Recall}&\textbf{F-score}&\textbf{FPS}\\
    \hline\hline
    Kang \textit{et al.}\cite{kang_orientation_2014} & 71 & 62 & 66 & - \\\hline
    Yao \textit{et al.}\cite{yao_detecting_2012} & 63 & 63 & 60 & 0.14 \\\hline
    Yin \textit{et al.}\cite{yin_multi-orientation_2015} & 81 & 63 & 74 & 0.71 \\\hline
    Yin \textit{et al.}\cite{yin_robust_2014} & 71 & 61 & 65 & 1.25 \\\hline
    Zhang \textit{et al.}\cite{zhang_multi-oriented_2016} & 83 & 67 & 74 & 0.48 \\\hline
    Yao \textit{et al.}\cite{yao_scene_2016} & 77 & \textbf{75} & 76 & $\sim$1.61 \\\hline
    Seglink \cite{shi_detecting_2017} & \textbf{86} & 70 & \textbf{77} & 8.9 \\\hline
    \textbf{ArbiText} & 78 & 72 & 75 & \textbf{12.1} \\\hline
    \end{tabular}
    \end{center}
\end{table}

\begin{figure}
\subcaptionbox{\label{sfig:testa}}{\includegraphics[width=2.6CM]{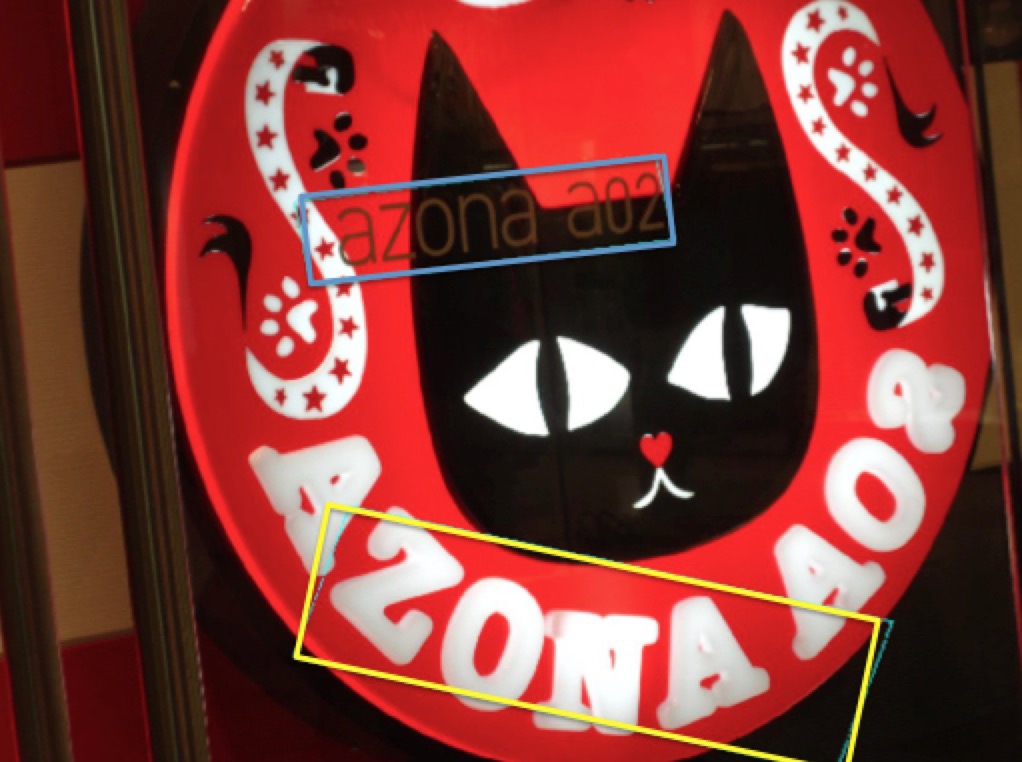}}
\subcaptionbox{\label{sfig:testa}}{   \includegraphics[width=2.6CM]{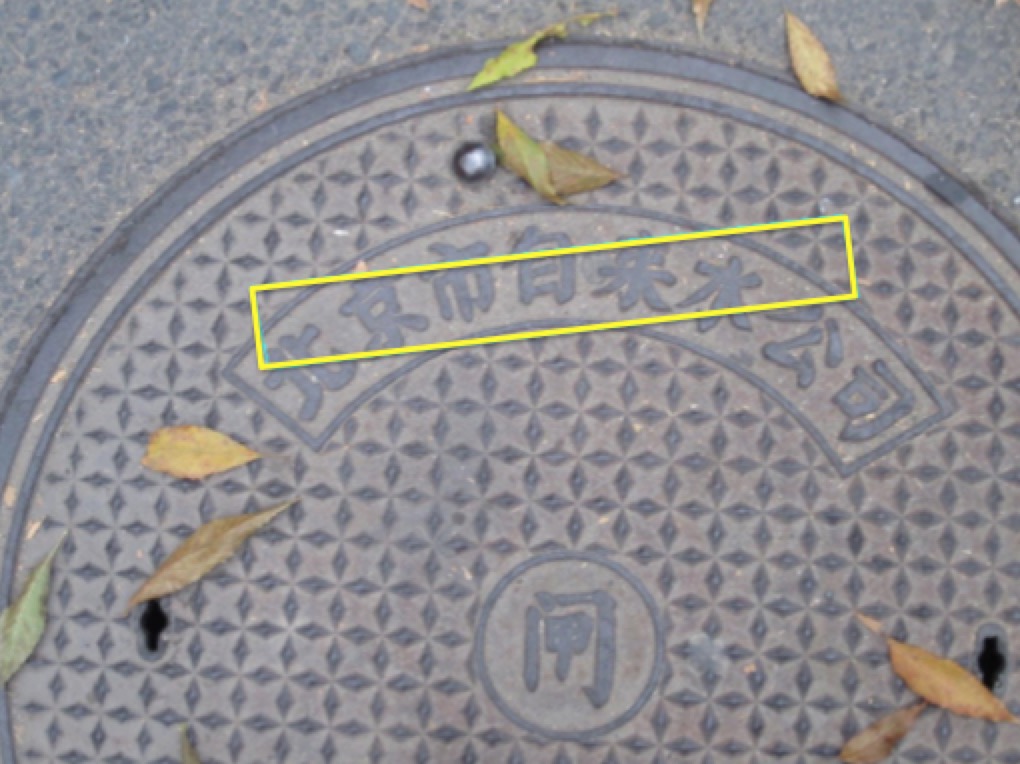}}
\subcaptionbox{\label{sfig:testa}}{\includegraphics[width=2.6CM]{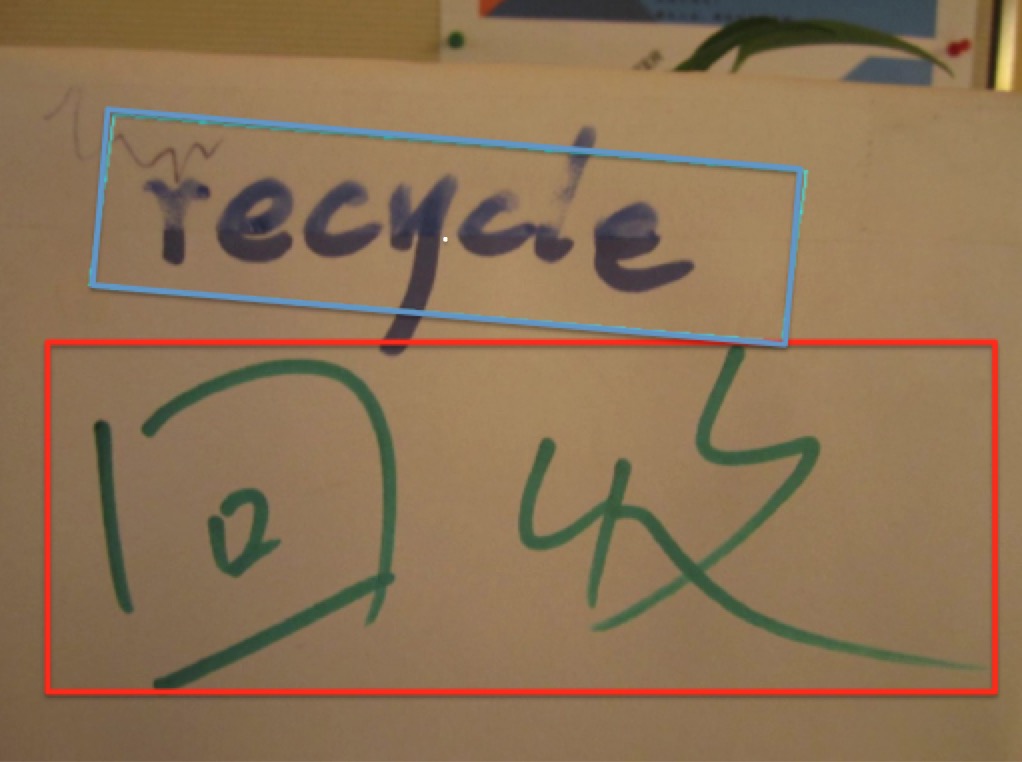}}
\caption{\textbf{Failure Cases On MSRA-TD500} The blue rectangles are true positives. The red ones are false negatives and yellow ones are false positives. In (a) and (b),  \textbf{ArbiText} fails to detect curved texts. In (c), \textbf{ArbiText} fails to detect certain hand-written texts}  
\label{fig:1}
\end{figure}

\subsection{Limitations}
As shown in Figure.9.a,b, curved texts can't be represented by circle anchors. Moreover, Figure.9.c shows our model's weakness in detecting hand-written texts.

\section{Conclusion}
We have presented ArbiText, a novel, proposal-free object detection method that can be utilized to detect both arbitrary-oriented texts and generic objects simultaneously. Its outstanding performance on different benchmarks demonstrates that ArbiText is accurate, robust, and flexible for real-world applications. In the future, we will extend the Circle Anchor methodology to detect deformable objects and/or texts.

{\small
\bibliographystyle{ieee}
\bibliography{Zotero}
}

\end{document}